\documentclass[letterpaper, 10 pt, conference]{ieeeconf}  

\IEEEoverridecommandlockouts                              

\usepackage{amsmath} 
\usepackage{amssymb}  
\usepackage{booktabs}
\usepackage{multirow}
\usepackage{graphicx}
\usepackage{subfigure}
\usepackage{cite}
\usepackage{float}

\usepackage{tikz}
\usetikzlibrary{positioning} 
\usetikzlibrary{matrix}
\usetikzlibrary{shapes}
\usepackage{pgfplots}
\pgfplotsset{compat=1.15}
\usepackage{pgfplotstable}

\usepackage{multirow, makecell}
\usepackage{array}

\usepackage[normalem]{ulem}
\useunder{\uline}{\ul}{}
\usepackage{siunitx}
\usepackage{tabularx}
\usepackage{tabu}
\PassOptionsToPackage{hyphens}{url}
\usepackage{hyperref}
\usepackage{censor}

\title{\LARGE \bf Likely, Light, and Accurate Context-Free \\ Clusters-based Trajectory Prediction}

\author{Tiago Rodrigues de Almeida and Oscar Martinez Mozos
\thanks{This work was supported by the Wallenberg AI, Autonomous Systems and Software Program (WASP) funded by the Knut and Alice Wallenberg Foundation.}
\thanks{The authors are with Center for Applied Autonomous Sensor Systems (AASS), \"{O}rebro University, \"{O}rebro, Sweden 
        {\tt\small \{tiago.almeida,oscar.mozos\}.@oru.se}}%
}

\begin{document}

\maketitle
\thispagestyle{empty}
\pagestyle{empty}

\begin{abstract}
Autonomous systems in the road transportation network require intelligent mechanisms that cope with uncertainty to 
foresee the future. In this paper, we propose a multi-stage probabilistic approach for trajectory forecasting: 
trajectory transformation to displacement space, clustering of displacement time series, trajectory proposals, and 
ranking proposals. We introduce a new deep feature clustering method, underlying self-conditioned GAN, which copes 
better with distribution shifts than traditional methods. 
Additionally, we propose novel distance-based ranking proposals to assign probabilities to the generated trajectories 
that are more efficient yet accurate than an auxiliary neural network. The overall system surpasses context-free deep
generative models in human and road agents trajectory data while performing similarly to point estimators when 
comparing the most probable trajectory.
\end{abstract}
\section{INTRODUCTION}
\label{sec:intro}

For a wide range of applications, an accurate estimation of the future position of moving agents is paramount. 
In the transportation sector, several branches benefit from effective forecasting systems such as (1) flexible, reliable and 
robust road traffic network relies on the prediction of traffic to prevent congestion and provide safety and efficiency~\cite{yu22,li23}; 
(2) Advanced Driver Assistance Systems (ADAS) capable of foreseeing the future can then adapt accordingly~\cite{singh22}; (3) the task of 
transporting goods may resort to a simulation of the future, which foster more informed planning and resource allocation~\cite{dalgkitsis21}. 
Further, trajectory prediction enhances situational awareness 
in complex dynamic environments such as industrial facilities, public sites, or home environments where multiple agents are moving. Finally, multimodal probabilistic forecasting of trajectories yields a richer 
representation of possible future events, which triggers subsequent decision-making systems~\cite{dahl23}.

The trajectory forecasting task mainly underlies three pillars such as agents interactions modeling~\cite{kosaraju19,sadeghian19,tang22,su22,minoura23,kothari23},
semantic context learning~\cite{xia22,kress23}, and multimodal predictions~\cite{deo22,bae22,chen22-itsc}. Also, multimodal approaches brought
a new view to the trajectory prediction domain as accounting for a set of diverse behaviors, instead of a unique point estimate, offers a much more
reasonable and safe solution~\cite{gupta18}. The problem with previous approaches is that they do not 
provide probabilistic or score measures for each trajectory estimate. In these cases, depending on the model's variability (induced usually by the
\emph{k-variety loss}~\cite{gupta18}), the predictions can be considered accurate or not~\cite{kothari23}. 
Most works following previous benchmarks~\cite{kothari21} opt by sampling $K$ trajectories (usually, $K=20$) and evaluating the closest to the ground truth. 
The premise that the closest trajectory to the ground truth can give a reasonable overview of the forecaster's performance is ambiguous as the model's 
variability plays a major role in the reported results~\cite{kothari23}. Therefore, the research community urges probabilistic and accurate predictions with 
adequate results reporting. In addition, context-agnostic 
approaches have had little attention from the research community~\cite{becker18,scholler20,giuliari20}, which we see as a gap to explore novel strategies 
for adding trajectory-based information to improve current context-free methods.

\begin{figure}[t]
    \centering
    \tikzset{myLine1/.style = {->, thick, >=stealth},
        mynarrownodes/.style = {node distance=0.75cm and 0.3cm},
    }
    \begin{tikzpicture}
        \tikzstyle{bigbox} = [draw=black!50, thick, fill=gray!50, rounded corners, rectangle]
        \tikzstyle{box} = [minimum size=0.2cm, rounded corners,rectangle split,rectangle split parts=2, inner sep=0.2ex, fill=gray!10]
        \node[inner sep=0, anchor=center] (input) at (0,0)
        {\includegraphics[width=0.40\columnwidth]{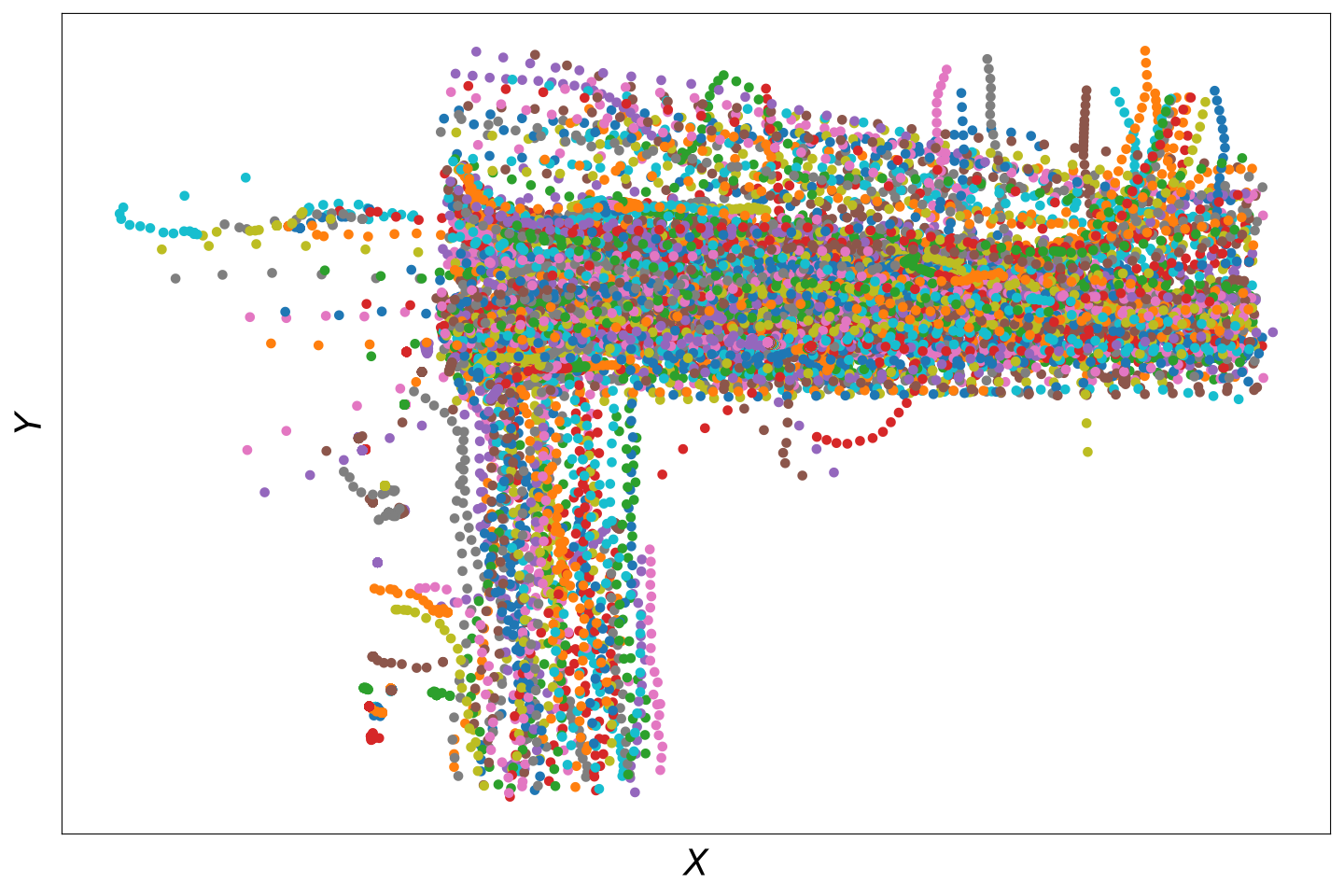}};
        \node[mynarrownodes,align=center,above=0.05mm of input](traj_str) {Trajectories};

        \node[inner sep=0pt,right=1cm of input] (input_disp)
        {\includegraphics[width=0.40\columnwidth]{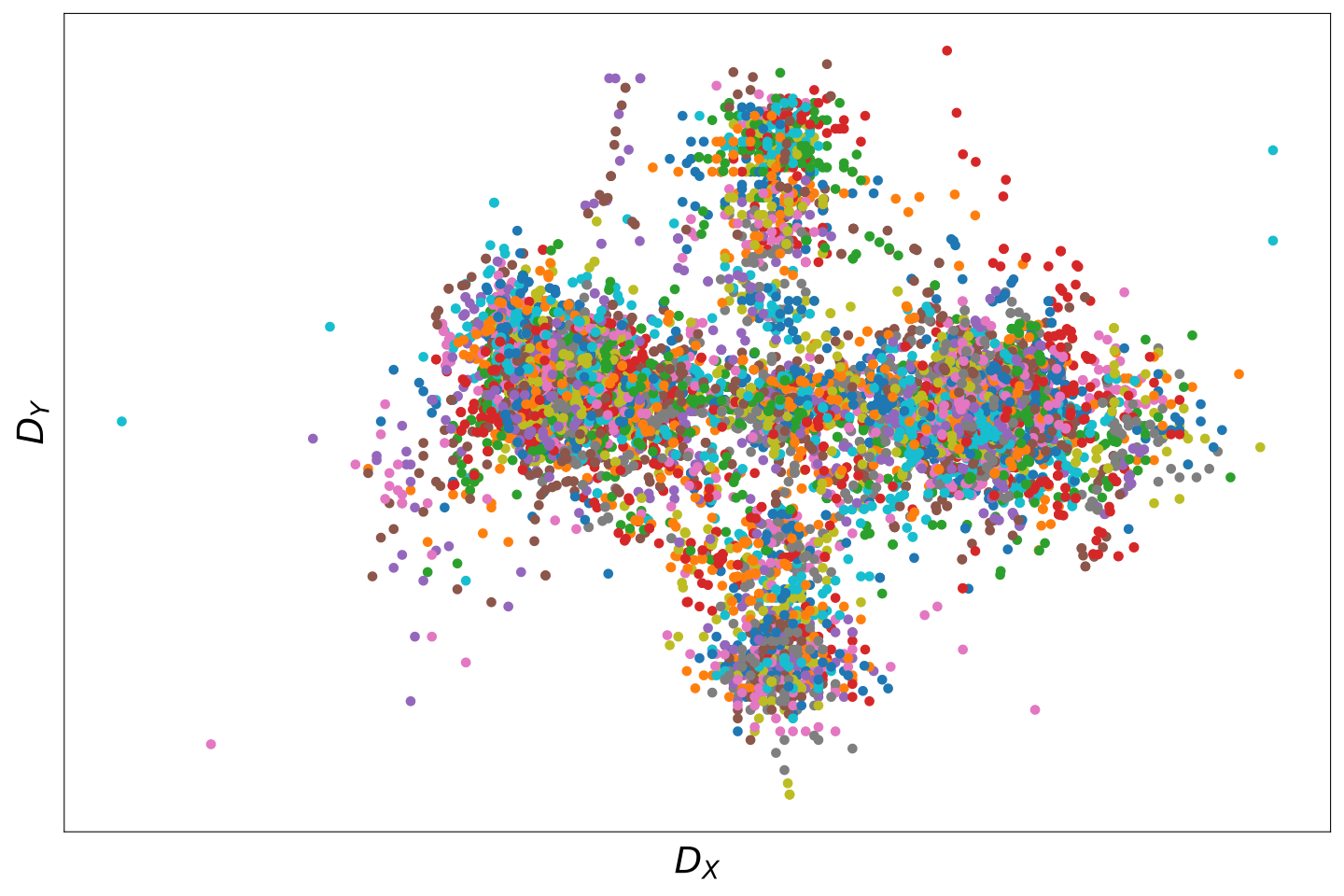}};
        \node[mynarrownodes,align=center,above=0.05mm of input_disp](disp_str) {Displacements};

        \node[inner sep=0pt,below=1.4cm of input_disp] (clusters)
        {\includegraphics[width=0.40\columnwidth]{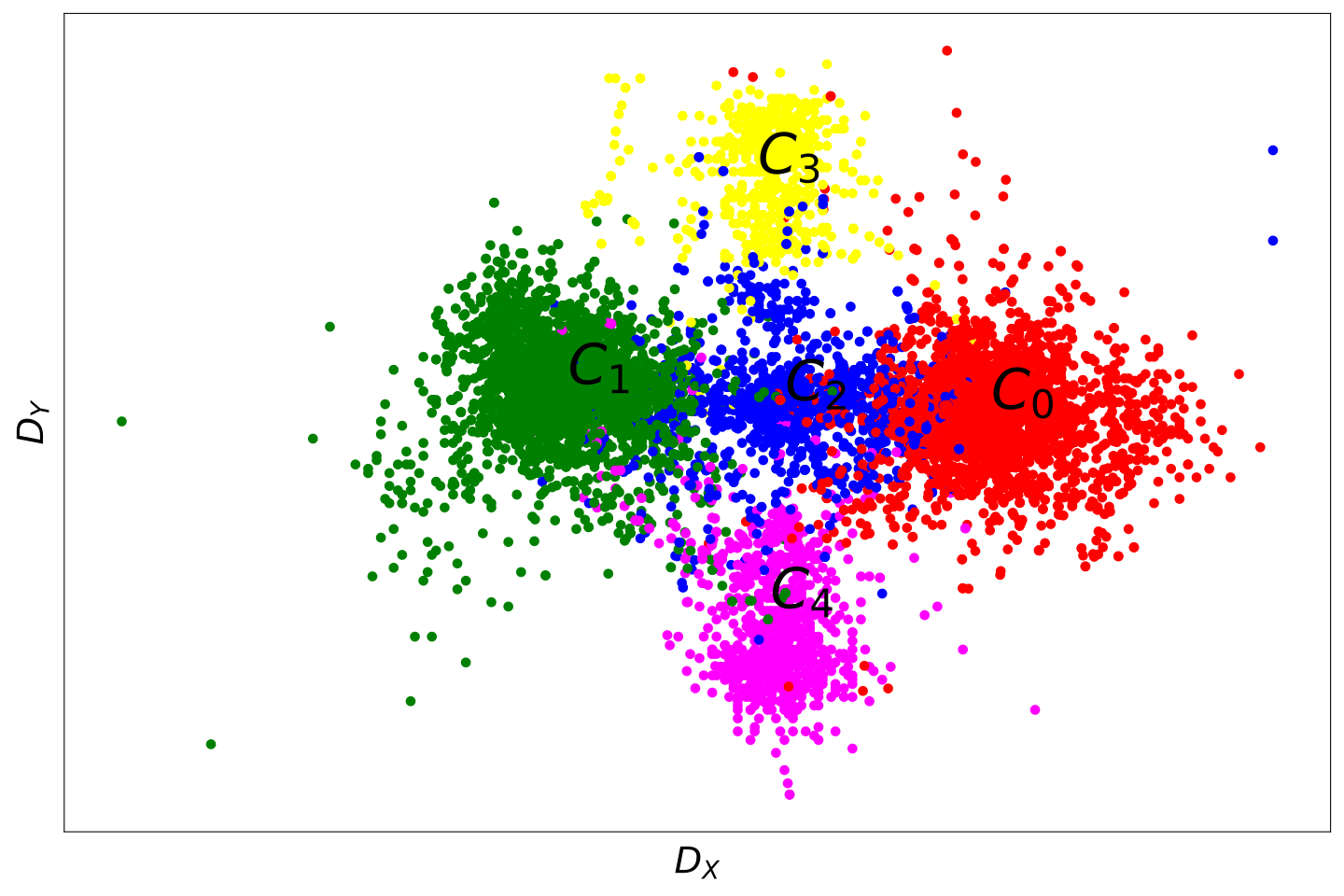}};
        \node[mynarrownodes,align=center,above=0.05mm of clusters](cl_str) {Clusters};

        \node[inner sep=0pt,below=0.7cm of input] (diag_train_inf)
        {\includegraphics[width=0.40\columnwidth]{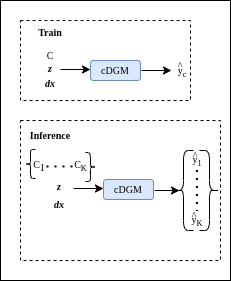}};
        \node[mynarrownodes,align=center,above=0.05mm of diag_train_inf](gen_pp) {Generative Proposals};
        \node[draw,rectangle,fill=gray!20,below=0.7cm of clusters] (rank) {Ranking Proposals};

        \draw[myLine1] (input.east)   to[out=0, in=180] (input_disp.west);
        \draw[myLine1] (input_disp.south)   to[out=-90, in=90] (cl_str.north);
        \draw[myLine1] (clusters.west)   to[out=180, in=0] (gen_pp.east);
        \draw[myLine1] (diag_train_inf.east)   to[out=0, in=180] (rank.west);

        \node at (2.25, 0.2) {1};
        \node at (4.7, -1.55) {2};
        \node at (2.45, -2.5) {3};
        \node at (2.6, -5) {4};
    
    \end{tikzpicture}
    \caption{System overview. (1) We transform spatiotemporal trajectory data into the displacement space. 
    (2) We cluster the displacement data into $K$ partitions. (3) We train deep generative models that take as 
    inputs the displacements ($dx$), the respective cluster class ($C$), and white noise ($z$). During inference, those models propose 
    $K$ predictions, being $K$ the number of clusters. 
    (4) We proceed to the ranking proposals step, where we assign probabilities to the predicted trajectories.}
    \label{fig:system_overview}
\end{figure}
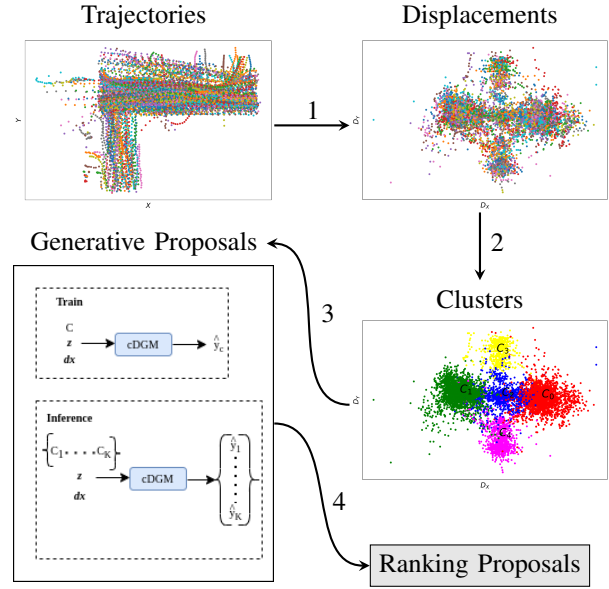

Following recent works~\cite{sun21,chen22,miao22}, our method encompasses multiple stages where the final objective is to provide a distribution of trajectories and the respective probabilities. To that end, our 
system sequentially clusters the input data, trains a conditional Deep Generative Model (cDGM) to map the input data and clusters' ids to the 
respective future trajectory, and assigns likelihoods to the trajectories in a post-hoc fashion (see Fig.~\ref{fig:system_overview}). The clusters ids 
aim to represent akin similar behaviors that drive the final predictions whereas the ranking proposals methods allow us to assign probabilities to each 
generated trajectory. In this way, we can provide diverse and probabilistic accurate predictions.
Furthermore, inspired by~\cite{kothari23}, we consider Top-3 scores for the assessment and comparison of methods. Previous works have considered the Top-20 
trajectories, which we agree can mislead the interpretation of the results.

In summary, this work aims to improve standalone context-free Deep Generative Models by adding trajectory-related information encoded in the conditioning cluster class.
Inspired by~\cite{liu20,tiago22}, our first contribution is a novel deep generative clustering algorithm, Full Path Self-Conditioned GAN (FP SC-GAN). 
Besides clustering the input data, this framework generates complete sets of displacements that can be part of a downstream data augmentation process.
However, in this work, we solely focus on the clustering capability of FP SC-GAN. 
Alternatively to~\cite{sun21,chen22,miao22}, 
our second contribution is the distance-based ranking proposals methods, which do not rely on training an auxiliary neural network but are still rather effective
and efficient.
The proposed distance-based ranking proposals methods uniquely require access to the clustering space. In this work, these methods perform likewise to an auxiliary deep neural network, and 
run in \emph{linear time} whereas a Multilayer Perceptron (MLP) requires \emph{quadratic time}. 
Finally, to the best of our knowledge, we propose the first quantitative metrics report where the likelihoods provided by the system directly affect the Top-3 trajectory prediction
metrics.

\section{RELATED WORK}
\label{sec:related_work}


Trajectory forecasting has been predominantly researched on two fronts: context-agnostic and context-aware approaches. 
Context-agnostic methods solely forecast based on observed trajectory patterns, whereas context-aware methods include social and scene layout cues. 
While context-agnostic approaches have received little attention~\cite{becker18,scholler20,giuliari20} from the research community, context-aware 
methods have been comprehensively investigated~\cite{gupta18,kosaraju19,dendorfer20,sun20,salzmann20,kothari23}. 
This paper focuses on context-agnostic approaches and breeding mechanisms for probabilistic trajectory forecasting based on encoded clustering 
information.

In the trajectory generation domain, some works tackled the problem from a probabilistic 
view in road scenarios~\cite{phan-minh20,ma21,ivanovic22,calem22} and human trajectory data~\cite{sun21,chen22,miao22}.
CoverNet~\cite{phan-minh20} comprises a Convolutional Neural Network (CNN) to extract contextual features from a road scene 
and a trajectory generator module to produce a set of possible predictions. Then, the system directly classifies the set of 
plausible trajectories yielding the score of each prediction. Conversely, we aim to produce context-agnostic samples to open 
the domain of applications of our system and reduce its requirements.
In~\cite{ma21}, the authors propose a post-hoc method named Likelihood-Based Diverse Sampling (LDS). In that paper, a novel objective function and a non-i.i.d sampling method encourage diverse predictions by suppressing similar predictions from the generated set of samples and leveraging the likelihoods from a pre-trained generative model. However, this work does not determine the scores (likelihoods) of the set of plausible trajectories, which we claim is paramount for risk-aware downstream
decision processes. 
\emph{HAICU}~\cite{ivanovic22} is a system that relies on perception and classification modules to give the 
class distribution of road agents. This system stands for a Conditional Variational Autoencoder (cVAE) conditioned on the class distribution and the observed trajectory of 
road agents to produce multimodal predictions. Therefore, this work heavily relies on upstream supervised methods to yield the class distribution, which is not easily generalizable to human trajectory data. To cope with this limitation, we propose unsupervised techniques to cluster akin trajectories, which we consider agnostic to the trajectory domain and more generalizable. 
In~\cite{calem22}, the authors enforce underlying physical admissibility constraints and diversity in a post-hoc trajectory sampling process 
based on a determinantal point process (DDP). Although it proposes a robust strategy for considering context using admissibility
constraints induced in the objective function, it does not provide a probabilistic view of the predicted trajectories.
\cite{sun21} devises a three-step method based on clustering, classification, and synthesis to predict. Contrarily to this work,
we first predict by using a cDGM and then propose a ranking proposals step. In this work, we investigate ranking proposals methods based on the distance 
to the already conceived clustering space and, therefore, not relying on a classification network. Further, \cite{chen22} 
proposes a method based on clustered goal points and a final classification step. While both \cite{sun21} and \cite{chen22}, 
at the classification step, learn the mapping between the past trajectory and the cluster class, we rank complete trajectories emphasizing the generated {\em tracklets}. Finally, in \cite{miao22}, 
the authors investigated a system with two branches: a motion pattern selector 
and a multimodal trajectory generator. The former produces a \emph{gallery} of diverse motion patterns, while the latter refines them and 
generates future trajectories. Then, a scoring method produces the most diverse predictions.

Our work encompasses mechanisms under the same umbrella as~\cite{sun21,chen22,miao22}, but our main objective is to propose a system that can 
improve current deep generative models by including information from clustered data. In our system
the clusters drive the multimodality, and the ranking proposals methods run on the generated trajectories. 
In addition, our distance-based ranking 
proposals methods does not rely on training any auxiliary neural network conversely to previous works~\cite{sun21,chen22,miao22}. In our case,
the ranking proposals step depends exclusively on the intrinsic nature of the clustering space and a similarity measure to the generated trajectories. Further, 
we propose a new deep clustering method inspired by a self-supervised deep generative model developed for the image generation task~\cite{liu20}. 
Finally, contrarily to previous works, the probabilities of each trajectory strongly affect the evaluation of our method, while in previous works, 
the probabilities only give a sense of the likelihood of each particular event.
\section{Methodology}
\label{sec:methodology}

The objective of our system (see Fig.~\ref*{fig:system_overview}) is to offer accurate and probabilistic future estimations for trajectory data. Succinctly, it comprises four sequential steps: 
(1) transformation to the displacement space, (2) clustering step, (3) deep generative proposals, and (4) ranking proposals step. 
To gain a comprehensive understanding of each component, this section provides a detailed description of the problem we address 
and an overview of our proposed solution. Subsequently, we delve into the displacement transformation and 
clustering steps. In the succeeding subsection, we outline the process of acquiring the proposals. Lastly, we introduce the ranking 
proposals step, which assigns probabilities to the set of predicted outputs.

\subsection{Problem Statement and System Overview} \label{sec:problem_satement}

Most trajectory forecasting models generate a set of plausible futures, $\{\boldsymbol{\hat{y_i}}\}^{K}$,  given an observed {\em tracklet}, 
$\boldsymbol{x}$.
The observed {\em tracklet} is an evenly spaced time series of $T_O$ observations in the  
$XY$\nobreakdash-plane. Ultimately, the forecasting models' goal is to predict the future behavior of the input time series 
for a certain horizon of $T_P$ time steps. In this paper, we go beyond the trajectory generative process and also propose a 
post-hoc method to assign probabilities, $\{\hat{p_i}\}^{K}$, to the respective $K$ proposed trajectories. 
In this way, besides future plausible trajectories we also provide the likelihood of each trajectory. We consider the system's 
starting point a dataset composed of $N_T$ trajectories.

In the first step of our system, we transform the input trajectories into the displacement space.
As a result, we also perform any prediction in the displacement space.
In doing so, we avoid dependence on the spatial context where the input data lies. Therefore,
any downstream task is more generalizable to new domains. Then, 
we cluster the displacement input vectors 
into $K$ partitions via clustering. This step is paramount
in the system as it groups akin time series of displacements that we feed into 
the generation process of future displacements. Consequently, we propose to train cDGMs, such as 
a cVAE or a Conditional Generative Adversarial Network (cGAN) to learn the future displacements given the observed displacements
and the respective cluster class, $c$ (i.e., the identifier for the 
cluster class conceived with the ground truth displacements). As we cluster the
entire set of displacements, after training the cDGM, 
there will be a link between the prediction and both the observed displacement 
{\em tracklet} and the respective cluster class. Assuming a successful clustering and cDGM's training steps,
this link allows us to claim that the prediction generated from the cluster class $c$ will be more similar to the group of trajectories
in $c$ than the trajectories from any other cluster. Thus, during inference, we propose to generate 
$K$ future displacement vectors (one per cluster class) and use a ranking method 
to assign likelihoods to the proposed future displacements. The idea behind the ranking method is 
to assign high likelihoods to displacement vectors that resemble the ones on the respective ground truth cluster 
and do the opposite for the remaining ones. For instance, assuming that the right cluster class is $c_{i}$, 
our key insight is that the produced sample for $c_i$, $\hat{\boldsymbol{dy_i}}$,   will be more 
similar to the ground truth displacements on $c_i$ than any prediction in $\{\boldsymbol{\hat{dy_j}}\}_{j=1 \setminus i}^K$
to the respective ground truth displacements in each of the clusters in 
$\{{c_j}\}_{j=1 \setminus {i}}^{K}$.

\subsection{Displacement Transformation and Clustering Step} \label{sec:clustering_step}

Primarily, our method transforms the raw trajectory data, $\boldsymbol{x}\oplus\boldsymbol{y}$, 
into displacement vectors, $\boldsymbol{D} \in {\rm I\!R}^{N_{T} \times (T_O + T_P)\times 2}$, 
where each displacement vector is given by $\boldsymbol{d}=\boldsymbol{dx}\oplus\boldsymbol{dy}$ or formally the finite differences of the
observed positions:
\begin{equation}
    s = \{(d_x, d_y)^1, ..., (d_x, d_y)^{T_P}\},
\label{eq:raw_s}
\end{equation}
being the superscript the time step. 

At the clustering step, we group $\boldsymbol{D}$ 
into sub-partitions of akin displacements, $\{ \boldsymbol{d_j} \}_{j=1}^{K}$, where $\boldsymbol{d_j}$ is a set itself with arbitrary
cardinality (less or equal than $N_{T}$). At the end of the clustering step, 
we guarantee that $\sum_{j=1}^{K} | \{\boldsymbol{d_j}\}| = N_T$, i.e., each sample belongs to only one
cluster. We evaluate three different clustering methods: \emph{k-Means}~\cite{jin10}, \emph{TS k-Means}~\cite{tslearn20}, and 
our proposed FP SC-GAN. 
To this end, we feed \emph{k-Means} with flattened displacements vectors in the form of:
\begin{equation}
    s_f = \{d_{x}^1, d_{y}^1, ..., d_x^{T_P}, d_y^{T_P}\}
\end{equation}

\begin{figure}[t]
    \centering
    \tikzset{myLine1/.style = {->, thick, >=stealth},
    mynarrownodes/.style = {node distance=0.75cm and 0.3cm},
    ball/.style={
        ellipse,
        minimum width=3cm,
        minimum height=1.5cm,
        draw
    },
    }
    
    \begin{tikzpicture}
        \tikzstyle{bigbox} = [draw=black!50, thick, fill=gray!50, rounded corners, rectangle]
        \tikzstyle{box} = [minimum size=0.2cm, rounded corners,rectangle split,rectangle split parts=4, inner sep=0.2ex, fill=gray!10]
        
        \node[inner sep=0, anchor=center] (input_disc_fake) at (0,0) {$\boldsymbol{\hat{dx}\oplus\hat{dy}}$};
        \node[align=center,below=0.50mm of input_disc_fake](or) {or};
        \node[align=center,below=0.05mm of or](input_disc_real) {$\boldsymbol{dx \oplus dy}$};

        \matrix[right=1.20cm of or, row sep=0mm, column sep=7mm, inner sep=2mm, bigbox, every node/.style=box](disc_box) {
            \node(encoder_box){\textbf{Encoder   } \vphantom{$\vcenter{\vspace{1.5em}}$} \nodepart{second} LSTM  \nodepart{third} or \nodepart{fourth} MLP}; 
            &  \node(classifier_box){\textbf{Classifier} \vphantom{$\vcenter{\vspace{1.5em}}$} \nodepart{second} \nodepart{third} MLP \nodepart{fourth}}; \\
        };
        \node[align=center, above=0.25mm of disc_box](disc_string) {DISCRIMINATOR ($D$)};
        \node[align=center, right=7cm of or](scores) {$s$};

        \draw[myLine1] (or.east) to[out=0, in=180] (disc_box.west);
        \draw[myLine1] (disc_box.east) to[out=0, in=180] (scores.west);
        \draw[myLine1] (encoder_box) -- (classifier_box) node [sloped,midway](M){};

        \node[ball, align=center,below=0.5cm of disc_box](clustering) {};
        
        \node[circle,draw,fill=gray!25, minimum size=0.1cm] at (4.65,-2.72) (m1) {};
        \node[ellipse,rotate=35,draw,fill=gray!25,minimum width=1.2cm,minimum height=0.5cm,above left=0.30cm of m1](m2) {};
        \node[ellipse,rotate=80,draw,fill=gray!25,minimum width=0.3cm,minimum height=1.0cm,left=0.30cm of m2](m3) {};
        \node[circle,draw,fill=gray!25, minimum size=0.40cm, above=0.2cm of m3](m4) {};
        \node[ellipse,draw,fill=gray!25,minimum width=0.2cm,minimum height=0.60cm,right=0.2cm of m4](m5) {};
        \node[ellipse,rotate=60,draw,fill=gray!25,minimum width=0.23cm,minimum height=0.64cm,below right=0.3cm of m3](m6) {};
        
        \draw[myLine1] (M) -- (clustering);
        
        \node[align=center,below=4.4cm of or](input_gen){$\boldsymbol{z}\oplus c$};

        \matrix[right=2.25cm of input_gen, row sep=0mm, column sep=10mm, inner sep=2mm, bigbox, every node/.style=box](gen_box) {
             \node(decoder_box){\textbf{Decoder} \vphantom{$\vcenter{\vspace{1.5em}}$} \nodepart{second} \nodepart{third} MLP \nodepart{fourth}}; \\
        };
        \node[align=center, above=0.25mm of gen_box](gen_string) {GENERATOR ($G$)};
        \node[align=center, right=1.25cm of gen_box](pred) {$\boldsymbol{\hat{dx}\oplus\hat{dy}}$};
        \node[align=center, left=0.25mm of clustering](cluster_str) {Clustering};
        \draw[myLine1] (input_gen.east) to[out=0, in=180] (gen_box.west);
        \draw[myLine1] (gen_box.east) to[out=0, in=180] (pred.west);
        \draw[myLine1] (m3.west) to[out=-180,   in=90] (input_gen.40);

    \end{tikzpicture}
    \caption{FP SC-GAN architecture. We pass random noise concatenated with the cluster class, drawn according 
    to the clustering space distribution, to the generator. The generator aims to produce a complete set of 
    displacements. Both generated and real samples pass through the discriminator, which learns to classify 
    them into real or fake. During training, we periodically recluster the discriminator's feature space.}
    \label{fig:self_conditioned_gan}
\end{figure}
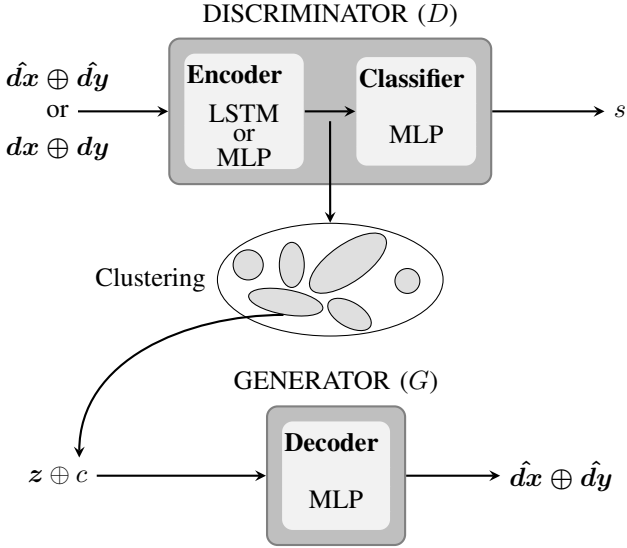

We also evaluate the extended version for time series provided by~\cite{tslearn20}, \emph{TS k-Means}. 
The major differences between these two clustering methods are that for the \emph{TS k-Means}, the input is given by Eq.~\ref{eq:raw_s} and the similarity 
measure is evaluated time-wise (we use \emph{soft-DTW}~\cite{cuturi17}).  
Finally, Fig.~\ref*{fig:self_conditioned_gan} depicts our proposed deep feature-based clustering method, FP SC-GAN. This architecture is considered a two-fold
provider: on one side, it produces the clustering space used in the later forecasting task; on the other side, it generates 
a complete set of synthetic displacements. This synthetic data can even serve as a data augmentation process, which we do not cover 
in this work. FP SC-GAN underlies the idea that the discriminator's feature space can be meaningful for downstream tasks~\cite{liu20} (in this case, the
clustering task). To that end, FP SC-GAN comprises a conditional generator ($G$) and a discriminator ($D$). We condition the generator on a cluster class
($c$) drawn from the clustering space and white noise ($\boldsymbol{z}$). Then one MLP-based decoder generates an entire displacement vector,
$\boldsymbol{\hat{d}}=\boldsymbol{\hat{dx}\oplus\hat{dy}}$, that should resemble one of the ground truth samples belonging to the conditioning cluster. The discriminator ($D$) learns 
to distinguish generated and ground truth samples according to a score ($s$).
During training, to assess the clustering 
process, we assume that the clustering is as good as the quality of the generated displacements. We consider this a 
fair \emph{proxy} as the only deterministic input signal given to the generator is the cluster id. 
Analogously to~\cite{tiago22}, we train the FP SC-GAN with the binary cross entropy loss
to optimize the discriminator and a weighted sum of the discriminator's loss and the mean squared error (MSE) loss to optimize the generator, given by:

\begin{equation}
\begin{aligned}
    L_{G} = & \lambda~MSE(\boldsymbol{y}, \boldsymbol{\hat{y}})\\
    & + (1 - \lambda)~(\frac{1}{2} \mathbb{E}[(D(\boldsymbol{d}) - 1)^2] 
    + \frac{1}{2} \mathbb{E}[D(\boldsymbol{\hat{d}})^2]),
\end{aligned}
\label{eq:gan_loss}
\end{equation}
being $\boldsymbol{\hat{y}}$ the entire generated $2D$ trajectory and $\boldsymbol{y}$ the respective 
ground-truth, $\boldsymbol{\hat{d}}$ the generated displacements vector and $\boldsymbol{d}$ the respective ground truth, 
and $\lambda$ the weight applied to the MSE term. For more details on the FP SC-GAN training, we refer the reader
to~\cite{tiago22,liu20}. 
%


\subsection{Deep Generative Proposals} \label{sec:deep_generative_proposals}


To produce potential displacement vectors, we train a cDGM that learns the mapping $\{\boldsymbol{dx}, c\}\rightarrow\boldsymbol{dy}$. 
In order to strengthen the validation of our results, we evaluate two cDGMs: a cVAE and a cGAN. In these methods, we first 
concatenate the cluster class to the input displacement vector. Then, we extract features ($f_e$) from this joint representation. The next 
step --- noise sampling --- is different in cVAE and cGAN. While training the former, a recognition netowrk, $q_\phi$, learns a low-dimensional latent representation ($\boldsymbol{z}$) by 
modeling the ground truth set of future displacements ($\boldsymbol{dy}$). During inference, it works like the conditional generator
in cGAN therefore, we sample 
this latent vector from a standard Gaussian Distribution. Afterward, we concatenate the latent representation to the temporal 
hidden representation learned by the feature extractor ($f_e$). Finally, we autoregressively generate
future displacements by decoding the last displacement vector and using $f_e$.

To optimize the cVAE, we use a weighted sum of a reconstruction loss, given by the MSE loss, and 
a regularization loss (Kullback-Liebler divergence) encouraging the learned distribution to match the prior distribution 

\begin{equation}
    \begin{aligned}
        L_V =  & \lambda~MSE(\boldsymbol{y}, \boldsymbol{\hat{y}})\\ 
    & - (1 - \lambda)~\beta~D_{KL}[q_\phi(\boldsymbol{z} | \boldsymbol{dy}, c) \rVert p(\boldsymbol{z} | \boldsymbol{dx}, c)],
    \end{aligned}
\end{equation}
where $\beta$ is the weight applied to the regularization loss, 
$q_\phi(\boldsymbol{z} | \boldsymbol{dy}, c)$ and $p(\boldsymbol{z} | \boldsymbol{dx}, c)$ are the recognition and predictor networks, respectively.
For the recognition network, we employ 
an LSTM followed by linear layers. The predictor, on the other hand, utilizes an MLP to embed the input, an LSTM to extract 
temporal features, and an MLP to generate the predictions.

To achieve optimal performance of the cGAN, we adopt Eq.~\ref{eq:gan_loss} for 
optimization. However, it should be noted that unlike the FP SC-GAN approach, the cGAN acts as a forecaster. Consequently, 
the inputs for the MSE and discriminator losses consist of future {\em tracklets}.

During inference, the trained cDGMs output a set of $K$ possible displacement vectors, $\{\boldsymbol{\hat{dy}_i}\}^K$. This set is the input of the last component in the system --- ranking proposals --- where we 
assign likelihoods to each item.

\subsection{Ranking Proposals Step} \label{sec:ranking_step}


The final step of the system aims to provide the mapping $\{\boldsymbol{\hat{dy}_i}\}^K\rightarrow\{p_i\}^{K}$, which means assigning 
probabilities to the respective $K$ predicted samples from each of the existing clusters. As mentioned before, the 
mapping to the probability space should ensure that samples from the right cluster have higher probabilities when compared to samples
from the remaining clusters.
To cope with this, one could train a deep neural network to learn the mapping $\{\boldsymbol{\hat{dy_i}}\}^K\rightarrow\{p_i\}^{K}$ 
by itself~\cite{sun21,chen22}. Alternatively, we propose mechanisms that rely on distance-based similarity measures: \emph{centroids} and \emph{neighbors}.
For both methods, we consider the inverse relationship between the distance and the similarity between samples. For instance, the smaller 
the distance to a cluster's centroid, the greater the similarity to the samples of that cluster. Similarly, in \emph{neighbors}, we hypothesize 
that the smaller the distance to $N_{neig}$ neighbors from the same cluster, the greater the similarity to those samples and so a 
higher probability of belonging to that cluster. Formally, in distance-based methods, the probability of prediction $\boldsymbol{\hat{dy_i}}$ 
belonging to the respective conditioning cluster, $c_i$, is given by:

\begin{equation}
    \hat{p}_{c_i} = \frac{\exp(\frac{1}{m_{c_i}} / \tau)}{\sum_{j}^K \exp(\frac{1}{m_{c_j}} / \tau)},     
\end{equation} 
which is the soft-argmax function over the inverse of the distances, $m$. For the \emph{centroids} approach, $m_{c_i}$ corresponds 
to the L2-distance to the conditioning cluster's centroid, $c_i$. For the \emph{neighbors} approach, $m_{c_i}$ is the average L2-distance 
to the $N_{neig}$ closest neighbors from cluster $c_i$. Formally, it is given as follows:
\begin{equation}
    m_{c_i} = \frac{1}{N_{neig}}\sum_j^{N_{neig}} L2(\boldsymbol{\hat{dy_i}}, \boldsymbol{dy_j})   
\end{equation}

We also evaluate a deep neural network~\cite{sun21,chen22}.
To this end, after training the cDGM, we use it to generate a dataset of displacements following the clustering space distribution. 
Then, we train a simple MLP that learns to classify the generated samples into the respective pseudo-labels 
given by the cluster assignments with the cross-entropy loss.

\section{EXPERIMENTS}
\label{sec:experiments}

In this section we compare our proposed methods and current baselines in different settings. 
Firstly, we describe the datasets, the baselines, and the assessment metrics. 
Secondly, we show and analyze quantitatively and qualitatively the results obtained in those datasets. 
Finally, we thoroughly analyze the ranking proposals methods.

\subsection{Datasets, Baselines, and Metrics}


We use two settings to evaluate our methods: train-test split and leave-one-dataset-out approach. For the former, we 
use two datasets: TH\"{O}R~\cite{rudenko20} and Argoverse~\cite{ming19}. TH\"{O}R is a human trajectory dataset, where the participants' roles --- {\em visitors}, {\em workers}, and {\em inspector}  --- are scripted. We prepocess the raw data as 
in~\cite{tiago22}, so we end up with $2052$ trajectories for training, $439$ for validation, and $441$ for testing.
Moreover, following the current benchmarks~\cite{kothari21}, we create
trajectories of $8$-time steps of observation (\SI{3.2}{s}) and $12$-time steps of forecasting (\SI{4.8}{s}). Argoverse is a 
road agents trajectory dataset, where there are also supervised classes:  autonomous vehicles ({\em av}), 
regular vehicles ({\em agents}), and other road agents ({\em others}). For this dataset, analogously to 
~\cite{tiago22},  we sample $5726$, $2100$, and $1678$ trajectories for training, validation, and testing sets, respectively. Additionally, 
we perform the leave-one-dataset-out in the widely used ETH/UCY benchmark~\cite{pellegrini09,leal14}. In this benchmark, 
similarly to TH\"{O}R, the observation length is 8-time steps (\SI{3.2}{s}) and the prediction length is  $12$-time steps (\SI{4.8}{s}).
This benchmark comprises five datasets --- ETH, HOTEL, UNIV, ZARA1, and ZARA2 --- of which four are used to train the model and 
the remaining is left for testing. We use no overlapping between segments of entire trajectories.


We conduct the experiments with baselines widely used in scientific works in the trajectory prediction field:

\begin{itemize}
    \item Constant Velocity Model (\emph{CVM})~\cite{scholler20} --- heuristic model that assumes that the humans move with constant velocity and direction. In this work we also include its comparison in road agents' data 
    (Argoverse). In~\cite{scholler20}, the velocity is given by the 
    projection of the last displacement but we use a weighted sum of the previous displacements based on a Gaussian kernel provided by \cite{rudenko21}. Both methods achieve similar results.  
    
    \item RED-LSTM predictor (\emph{RED})~\cite{becker18} --- it is a stack of an LSTM ($64$ hidden dimensions) 
    and a 
    2-layer MLP (hidden dimensions in $\{32, 16 \}$) that receives linearly embedded displacements (linear layer with $16$ hidden dimensions).
    After every layer, we use a PreLU activation function.

    \item Context-free GAN (\emph{CF-GAN}) and Context-free VAE (\emph{CF-VAE})~\cite{kothari21} --- deep generative approaches based on~\cite{kothari21} but we remove any mechanisms that aim to model social interactions since 
    we are solely interested in seeing the potential of trajectory-related additional information. \emph{CF-VAE}, \emph{CF-GAN}, and 
    our deep generative proposals methods described in Sec.~\ref{sec:deep_generative_proposals} have the same design choice: initial linear
    layer with $16$ hidden dimensions, an LSTM with $64$ hidden dimensions and a final linear layer with $32$ hidden dimensions to decode the temporal features. 
    To investigate the ability to generate a wide range of plausible trajectories, we incorporate the \emph{k-variety loss} 
    introduced in \cite{gupta18}, as a replacement for the MSE partial loss utilized in the generator's loss 
    (in the case of \emph{CF-GAN}) and the VAE's loss (in the case of \emph{CF-VAE}).

\end{itemize}

As described in Section~\ref{sec:deep_generative_proposals}, we compare two cDGMs: cVAE (\emph{OURS-VAE}) and cGAN (\emph{OURS-GAN}). 
Additionally, we evaluate three clustering algorithms: \emph{k-Means}, \emph{TS k-Means}, and the proposed \emph{FP SC-GAN}. Furthermore,
we have three ranking proposals methods: (1) based on the Euclidean distance to the centroids (\emph{cent}); (2) based on the 
Euclidean distance to the $N_{neig}$ closest neighbors (set to $20$) in the displacement space and feature space, denoted as \emph{neigh-ds} and \emph{neigh-fs}, 
respectively. It is important to note here that \emph{neigh-fs} is only used in \emph{FP SC-GAN} as it is the unique method that 
makes use of a deep feature space; (3) based on the classification provided by the auxiliary network (\emph{anet}).
As our methods rely on a clustering process, we need to determine the number of clusters in each dataset. To 
do so, we average the results of five runs of each clustering method. For the remaining experiments, we use the number 
of clusters that yields the smallest Davies–Bouldin Index (DBI)~\cite{davies79}.

The metrics we use to compare the different methods rely on Average Displacement Error (ADE) that measures the average Euclidean distance between the predicted positions 
and the ground truth and the Final Displacement Error (FDE) that measures the Euclidean distance between the final predicted position and the respective ground truth (at $t=T_P$).
To compare both deterministic point estimate predictors (\emph{CVM} and \emph{RED}) to stochastic multimodal models (\emph{CF-GAN}, 
\emph{CF-VAE}, \emph{VAE-OURS}, and \emph{GAN-OURS}), we provide the following metrics:
\begin{enumerate}
  \item Top-$3$ ADE/FDE (in meters) --- used to assess the multimodal estimates produced by generative models. From \emph{CF-GAN} and \emph{CF-VAE}, we sample $3$ trajectories and evaluate the closest one to the ground truth, whereas from \emph{GAN-OURS} and \emph{VAE-OURS}, we take the $3$ most likely trajectories and compare the closest one to the ground truth.
  \item Top-$1$ ADE/FDE (in meters) --- used to compare point estimate predictors to stochastic multimodal models. From \emph{CF-GAN} and \emph{CF-VAE}, we use the first prediction generated by the models, while from \emph{GAN-OURS} and \emph{VAE-OURS}, we use the most likely trajectory.
\end{enumerate} Finally, to compare the ranking proposals methods, we use accuracy (in \%). In this case we compare the output of the ranking proposals methods to the soft labels yielded by 
each clustering method.

\subsection{Results}

In this section we show the results obtained by our methods and the baselines in the datasets, where bold scores denote the best score, and the results from Deep Learning (DL)-based approaches are averaged over five runs. 
First, as the main focus of our approach is to generate a probabilistic yet accurate set of predictions, we depict in Tab.~\ref{tab:res1_adefde} 
the Top-3 ADE/FDE results for deep generative approaches. 
Our methodology exhibits superior performance compared to the baseline models across all datasets, indicating that among the 
three most probable predicted trajectories, our system produces a more accurate prediction. 
This finding reinforces the notion that generating multimodality within the conditioning cluster, as opposed to 
relying on the latent space learned through the \emph{k-variety loss}, leads to improved outcomes. 
Furthermore, it is noteworthy to mention that the various clustering methods and ranking proposal 
mechanisms yield similar results across the datasets. However, in some specific datasets, one 
can find better results in other configurations: in TH\"{O}R with \emph{GAN-OURS}, \emph{cent} yields $0.45\pm0.02$ and $0.75\pm0.03$ and in ZARA2 
where \emph{TS k-Means} followed by \emph{neig} yields $0.37\pm0.01$ and $0.62\pm0.01$, for ADE and FDE, respectively.

\begin{table}[t]
  \begin{minipage}{\columnwidth}
  \centering
  \caption{Top-3 ADE/FDE ($\downarrow$) metrics in the test sets.}
  \label{tab:res1_adefde}
  \resizebox{0.98\columnwidth}{!}{%
  \begin{tabular}{@{}c|cc|cc@{}}
  \toprule
  Datasets &
  \emph{CF-GAN} &
  \emph{GAN-OURS}$^1$ &
  \emph{CF-VAE} &
  \emph{VAE-OURS}$^1$ \\ \midrule
  TH\"{O}R &
    \begin{tabular}[c]{@{}c@{}}$0.57\pm0.01$\\ $1.04\pm0.03$\end{tabular} &
    \begin{tabular}[c]{@{}c@{}}$\mathbf{0.53\pm0.01}$\\ $\mathbf{0.84\pm0.03}$\end{tabular} &
    \begin{tabular}[c]{@{}c@{}}$0.62\pm0.02$\\ $1.05\pm0.05$\end{tabular} &
    \begin{tabular}[c]{@{}c@{}}$\mathbf{0.56\pm0.01}$\\ $\mathbf{0.89\pm0.02}$\end{tabular} \\ \midrule
  Argoverse &
    \begin{tabular}[c]{@{}c@{}}$1.62\pm0.07$\\ $2.81\pm0.14$\end{tabular} &
    \begin{tabular}[c]{@{}c@{}}$\mathbf{1.56\pm0.02}$\\ $\mathbf{2.69\pm0.02}$\end{tabular} &
    \begin{tabular}[c]{@{}c@{}}$1.96\pm0.02$\\ $3.44\pm0.06$\end{tabular} &
    \begin{tabular}[c]{@{}c@{}}$\mathbf{1.62\pm0.02}$\\ $\mathbf{2.82\pm0.04}$\end{tabular} \\ \specialrule{.2em}{.1em}{.1em}  
  ETH &
    \begin{tabular}[c]{@{}c@{}}$0.84\pm0.03$\\ $1.64\pm0.06$\end{tabular} &
    \begin{tabular}[c]{@{}c@{}}$\mathbf{0.77\pm0.04}$\\ $\mathbf{1.60\pm0.11}$\end{tabular} &
    \begin{tabular}[c]{@{}c@{}}$0.94\pm0.02$\\ $1.84\pm0.04$\end{tabular} &
    \begin{tabular}[c]{@{}c@{}}$\mathbf{0.82\pm0.02}$\\ $\mathbf{1.60\pm0.04}$\end{tabular} \\ \midrule
  HOTEL &
    \begin{tabular}[c]{@{}c@{}}$0.87\pm0.07$\\ $1.64\pm0.12$\end{tabular} &
    \begin{tabular}[c]{@{}c@{}}$\mathbf{0.81\pm0.10}$\\ $\mathbf{1.46\pm0.11}$\end{tabular} &
    \begin{tabular}[c]{@{}c@{}}$1.08\pm0.04$\\ $1.95\pm0.06$\end{tabular} &
    \begin{tabular}[c]{@{}c@{}}$\mathbf{0.97\pm0.08}$\\ $\mathbf{1.72\pm0.14}$\end{tabular} \\ \midrule
  UNIV &
    \begin{tabular}[c]{@{}c@{}}$0.56\pm0.01$\\ $1.08\pm0.02$\end{tabular} &
    \begin{tabular}[c]{@{}c@{}}$\mathbf{0.51\pm0.01}$\\ $\mathbf{0.98\pm0.03}$\end{tabular} &
    \begin{tabular}[c]{@{}c@{}}$0.61\pm0.01$\\ $1.16\pm0.01$\end{tabular} &
    \begin{tabular}[c]{@{}c@{}}$\mathbf{0.56\pm0.01}$\\ $\mathbf{1.06\pm0.02}$\end{tabular} \\ \midrule
  ZARA1 &
    \begin{tabular}[c]{@{}c@{}}$0.43\pm0.02$\\ $0.82\pm0.07$\end{tabular} &
    \begin{tabular}[c]{@{}c@{}}$\mathbf{0.37\pm0.01}$\\ $\mathbf{0.72\pm0.03}$\end{tabular} &
    \begin{tabular}[c]{@{}c@{}}$0.48\pm0.01$\\ $0.98\pm0.04$\end{tabular} &
    \begin{tabular}[c]{@{}c@{}}$\mathbf{0.44\pm0.01}$\\ $\mathbf{0.91\pm0.03}$\end{tabular} \\ \midrule
  ZARA2 &
    \begin{tabular}[c]{@{}c@{}}$0.46\pm0.01$\\ $0.81\pm0.05$\end{tabular} &
    \begin{tabular}[c]{@{}c@{}}$\mathbf{0.40\pm0.01}$\\ $\mathbf{0.65\pm0.03}$\end{tabular} &
    \begin{tabular}[c]{@{}c@{}}$0.49\pm0.01$\\ $0.86\pm0.05$\end{tabular} &
    \begin{tabular}[c]{@{}c@{}}$\mathbf{0.43\pm0.01}$\\ $\mathbf{0.73\pm0.01}$\end{tabular} \\ \bottomrule
  \end{tabular}%
  }
  \vspace{0.2em} 
  \\ \textsuperscript{1}\tiny{\emph{FP SC-GAN} + \emph{neig-fs}}
  \end{minipage}
  \end{table}

The fact that our system performs better on the leave-one-dataset-out setting suggests that the clusters are 
unbiased to the reference dataset. In particular, Tab.~\ref{tab:res_hotel_adefde} shows the results in the 
HOTEL dataset with \emph{GAN-OURS}, where the trajectories go in a different direction than most of the ones in the training 
set~\cite{scholler20}. Here we can see that the proposed \emph{FP SC-GAN} is more robust to the distribution
shift present in this dataset.

  \begin{table}[t]
  \begin{minipage}{\columnwidth}
  \centering
  \caption{Top-3 ADE/FDE ($\downarrow$) metrics in HOTEL with GAN-OURS.}
  \label{tab:res_hotel_adefde}
  \resizebox{0.8\columnwidth}{!}{%
  \begin{tabular}{c|lll}
  \toprule
  \begin{tabular}[c]{@{}c@{}}Ranking\\ proposals\end{tabular} &
    \multicolumn{1}{c}{\emph{K-means}} &
    \multicolumn{1}{c}{\emph{TS K-means}} &
    \multicolumn{1}{c}{\emph{FP SC-GAN}} \\ \hline
  \emph{cent} &
    \begin{tabular}[c]{@{}l@{}}$1.03\pm0.05$\\ $1.80\pm0.09$\end{tabular} &
    \begin{tabular}[c]{@{}l@{}}$1.04\pm0.05$\\ $1.81\pm0.06$\end{tabular} &
    \begin{tabular}[c]{@{}l@{}}$\mathbf{0.80\pm0.10}$\\ $\mathbf{1.44\pm0.12}$\end{tabular} \\ \hline
  \emph{neig}$^1$ &
    \begin{tabular}[c]{@{}l@{}}$0.98\pm0.04$\\ $1.74\pm0.06$\end{tabular} &
    \begin{tabular}[c]{@{}l@{}}$0.97\pm0.11$\\ $1.71\pm0.16$\end{tabular} &
    \begin{tabular}[c]{@{}l@{}}$\mathbf{0.81\pm0.10}$\\ $\mathbf{1.46\pm0.11}$\end{tabular} \\ \hline
  \emph{anet} &
    \begin{tabular}[c]{@{}l@{}}$1.02\pm0.06$\\ $1.78\pm0.10$\end{tabular} &
    \begin{tabular}[c]{@{}l@{}}$0.97\pm0.04$\\ $1.70\pm0.04$\end{tabular} &
    \begin{tabular}[c]{@{}l@{}}$\mathbf{0.90\pm0.06}$\\ $\mathbf{1.59\pm0.08}$\end{tabular} \\ \bottomrule
  \end{tabular}%
  }
  \vspace{0.2em} 
  \\ \textsuperscript{1}\tiny{\emph{neigh-ds} for \emph{k-Means} and \emph{TS K-means}; \emph{neigh-fs} for \emph{FP SC-GAN}}
\end{minipage}
\end{table}

To assess the variability of each model and the fine-grained accuracy, we also provide in 
Tab.~\ref{tab:res2_t1adefde} the Top-1 ADE/FDE scores. Apart from HOTEL, this result shows that our method yields similar Top-1 predictions 
to point estimate and multimodal baselines, where \emph{RED} stood out the most. 
Nevertheless, when considering a comparison between the generative models \emph{CF-GAN} and \emph{CF-VAE} and our proposed methods \emph{GAN-OURS} 
and \emph{VAE-OURS}, which share the same network structure, our methods demonstrate superior performance in terms of Top-3 scores across 
all datasets. Additionally, our methods achieve equal or superior Top-1 scores in all datasets. Notably, our methods also effectively alleviate the performance disparity resulting from the models' variability when comparing Top-3 and Top-1 results.
On top of that, our methods yield more information encoded in the probabilities 
assigned to the predicted trajectories, thus, creating a space of probable future locations rather than uninformative predictions.
It is also interesting to note that 
\emph{CVM} provides similar or better performance in human trajectory 
data. However, for road agents' trajectory data, this method cannot cope with the speed variation in the trajectories.

\begin{table}[t]
  \begin{minipage}{\columnwidth}
  \centering
  \caption{Top-1 ADE/FDE ($\downarrow$) metrics in the test sets.}
  \label{tab:res2_t1adefde}
  \resizebox{0.98\columnwidth}{!}{%
  \begin{tabular}{@{}c|cc|cc|cc@{}}
  \toprule
  Datasets &
    \emph{CVM} &
    \emph{RED} &
    \emph{CF-GAN} &
    \emph{GAN-OURS}$^1$ &
    \emph{VAE} &
    \emph{VAE-OURS}$^1$ \\ \midrule
  TH\"{O}R &
    \begin{tabular}[c]{@{}c@{}}$0.79$\\ $1.28$\end{tabular} &
    \begin{tabular}[c]{@{}c@{}}$\mathbf{0.65\pm0.01}$\\ $\mathbf{1.06\pm0.01}$\end{tabular} &
    \begin{tabular}[c]{@{}c@{}}$0.85\pm0.09$\\ $1.68\pm0.20$\end{tabular} &
    \begin{tabular}[c]{@{}c@{}}$0.76\pm0.02$\\ $1.31\pm0.05$\end{tabular} &
    \begin{tabular}[c]{@{}c@{}}$0.71\pm0.02$\\ $1.30\pm0.04$\end{tabular} &
    \begin{tabular}[c]{@{}c@{}}$0.71\pm0.02$\\ $1.21\pm0.04$\end{tabular} \\ \midrule
  Argoverse &
    \begin{tabular}[c]{@{}c@{}}$2.57$\\ $3.93$\end{tabular} &
    \begin{tabular}[c]{@{}c@{}}$\mathbf{1.84\pm0.01}$\\ $\mathbf{3.18\pm0.02}$\end{tabular} &
    \begin{tabular}[c]{@{}c@{}}$2.41\pm0.09$\\ $4.53\pm0.17$\end{tabular} &
    \begin{tabular}[c]{@{}c@{}}$1.92\pm0.02$\\ $3.27\pm0.03$\end{tabular} &
    \begin{tabular}[c]{@{}c@{}}$2.69\pm0.02$\\ $4.94\pm0.02$\end{tabular} &
    \begin{tabular}[c]{@{}c@{}}$1.95\pm0.01$\\ $3.39\pm0.04$\end{tabular} \\ \midrule
  ETH &
    \begin{tabular}[c]{@{}c@{}}$\mathbf{0.95}$\\ $2.11$\end{tabular} &
    \begin{tabular}[c]{@{}c@{}}$\mathbf{0.95\pm0.01}$\\ $1.94\pm0.02$\end{tabular} &
    \begin{tabular}[c]{@{}c@{}}$1.08\pm0.08$\\ $2.18\pm0.19$\end{tabular} &
    \begin{tabular}[c]{@{}c@{}}$0.96\pm0.03$\\ $1.99\pm0.05$\end{tabular} &
    \begin{tabular}[c]{@{}c@{}}$1.08\pm0.03$\\ $2.15\pm0.08$\end{tabular} &
    \begin{tabular}[c]{@{}c@{}}$\mathbf{0.95\pm0.02}$\\ $1.95\pm0.03$\end{tabular} \\ \midrule
  HOTEL &
    \begin{tabular}[c]{@{}c@{}}$\mathbf{0.42}$\\ $\mathbf{0.74}$\end{tabular} &
    \begin{tabular}[c]{@{}c@{}}$1.03\pm0.04$\\ $1.84\pm0.06$\end{tabular} &
    \begin{tabular}[c]{@{}c@{}}$1.03\pm0.05$\\ $1.99\pm0.11$\end{tabular} &
    \begin{tabular}[c]{@{}c@{}}$0.95\pm0.13$\\ $1.72\pm0.19$\end{tabular} &
    \begin{tabular}[c]{@{}c@{}}$1.18\pm0.05$\\ $2.21\pm0.08$\end{tabular} &
    \begin{tabular}[c]{@{}c@{}}$1.09\pm0.12$\\ $1.96\pm0.18$\end{tabular} \\ \midrule
  UNIV &
    \begin{tabular}[c]{@{}c@{}}$\mathbf{0.65}$\\ $1.29$\end{tabular} &
    \begin{tabular}[c]{@{}c@{}}$\mathbf{0.65\pm0.01}$\\ $\mathbf{1.27\pm0.01}$\end{tabular} &
    \begin{tabular}[c]{@{}c@{}}$0.74\pm0.03$\\ $1.49\pm0.05$\end{tabular} &
    \begin{tabular}[c]{@{}c@{}}$0.74\pm0.01$\\ $1.44\pm0.04$\end{tabular} &
    \begin{tabular}[c]{@{}c@{}}$0.71\pm0.01$\\ $1.41\pm0.01$\end{tabular} &
    \begin{tabular}[c]{@{}c@{}}$0.69\pm0.01$\\ $1.36\pm0.01$\end{tabular} \\ \midrule
  ZARA1 &
    \begin{tabular}[c]{@{}c@{}}$0.54$\\ $1.05$\end{tabular} &
    \begin{tabular}[c]{@{}c@{}}$\mathbf{0.45\pm0.01}$\\ $\mathbf{0.87\pm0.01}$\end{tabular} &
    \begin{tabular}[c]{@{}c@{}}$0.60\pm0.11$\\ $1.21\pm0.24$\end{tabular} &
    \begin{tabular}[c]{@{}c@{}}$0.47\pm0.01$\\ $0.92\pm0.03$\end{tabular} &
    \begin{tabular}[c]{@{}c@{}}$0.64\pm0.01$\\ $1.33\pm0.03$\end{tabular} &
    \begin{tabular}[c]{@{}c@{}}$0.53\pm0.01$\\ $1.10\pm0.02$\end{tabular} \\ \midrule
  ZARA2 &
    \begin{tabular}[c]{@{}c@{}}$0.55$\\ $0.91$\end{tabular} &
    \begin{tabular}[c]{@{}c@{}}$\mathbf{0.50\pm0.01}$\\ $\mathbf{0.80\pm0.01}$\end{tabular} &
    \begin{tabular}[c]{@{}c@{}}$0.65\pm0.02$\\ $1.24\pm0.09$\end{tabular} &
    \begin{tabular}[c]{@{}c@{}}$0.52\pm0.01$\\ $0.86\pm0.04$\end{tabular} &
    \begin{tabular}[c]{@{}c@{}}$0.60\pm0.02$\\ $1.13\pm0.07$\end{tabular} &
    \begin{tabular}[c]{@{}c@{}}$0.54\pm0.01$\\ $0.93\pm0.01$\end{tabular} \\ \bottomrule
  \end{tabular}%
  }
  \vspace{0.2em} 
  \\ \textsuperscript{1}\tiny{\emph{FP SC-GAN} + \emph{neig-fs}}
  \end{minipage}
  \end{table}

\begin{figure}[t]
  \centering
  \includegraphics[width=\linewidth]{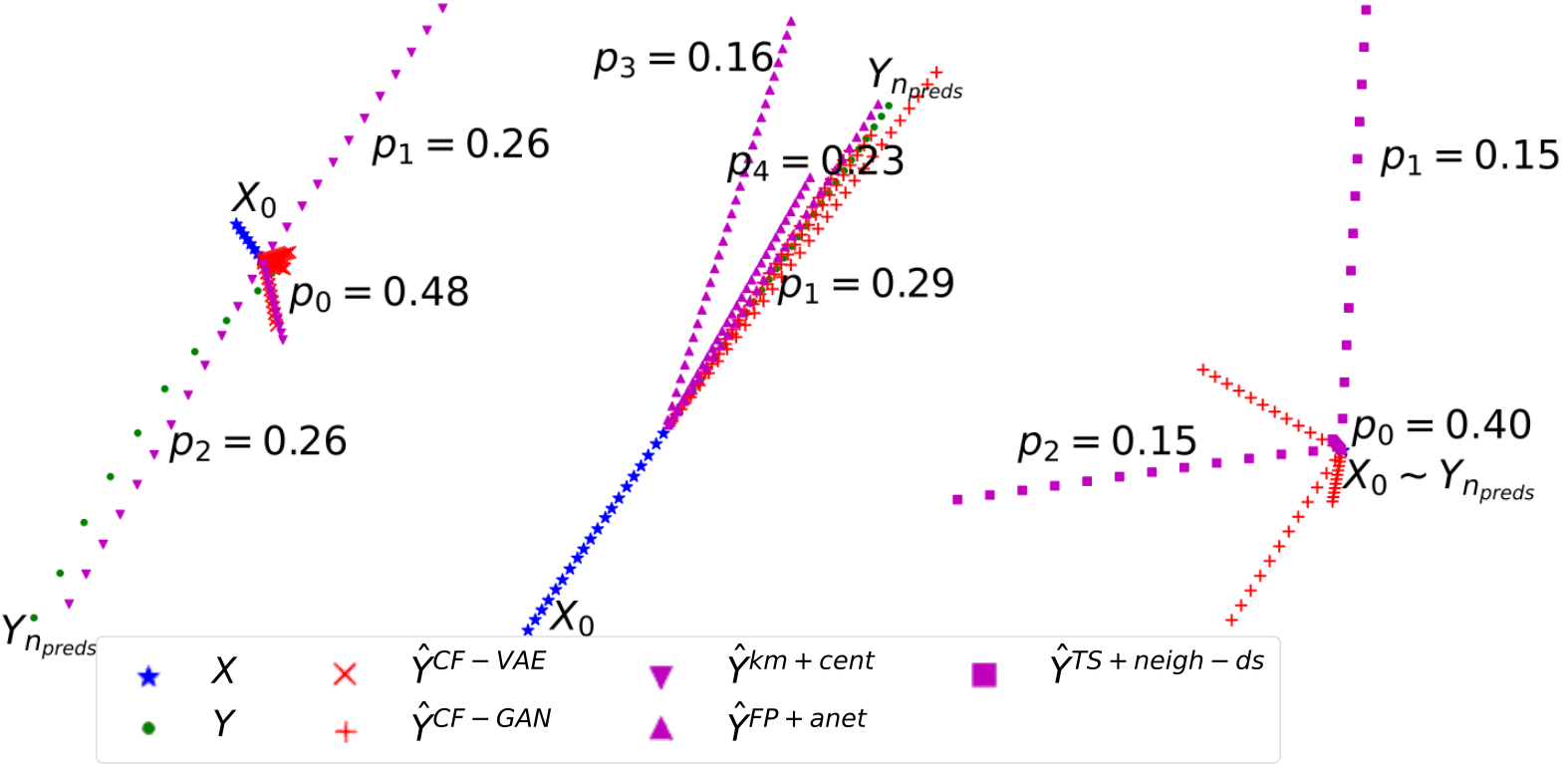} 
  \caption{Top-3 predictions in test samples from TH\"{O}R (left), Argoverse (center), and ZARA2 
  (right). $X_0$ denotes the first observed point, $Y_{n_{pred}}$ denotes the last point of 
  the ground-truth, and $p_{i\in[0,K-1]}$ the probabilities provided by our ranking proposals methods.}
  \label{fig:top3}
\end{figure}

In Fig.~\ref{fig:top3}, we show qualitative results of the Top-3 predictions from our 
methods and baselines in TH\"{O}R (left), Argoverse (center), and ZARA1 
(right) test sets. TH\"{O}R example pertains to a quite challenging scenario as the heading change 
is sharp. Still, our method could capture this uncommon behavior within 
the three most probable trajectories (with $p=0.26$), while the most probable trajectory ($p=0.48$) 
is following the movement's trend, which is reasonable due to the most common constant velocity profile in humans walking~\cite{scholler20}. In Argoverse 
example, the auxiliary network plays an important role in the hierarchical predictions: Top-1 prediction
($p_1=0.29$) is also the closest to the ground truth; the second most likely trajectory ($p_4=0.23$) 
is following the same direction but with a shorter distance; 
finally, the third most likely trajectory ($p=0.16$) goes in a different direction, but we consider it 
a still reasonable prediction. Finally, in ZARA2 example, while \emph{CF-GAN} could not capture the 
static behavior ($X_0 \sim Y_{n_{pred}}$) but still yields a broad range of 
behaviors, our method conditioned by the cluster id assigns the highest likelihood to the predicted 
static behavior ($p=0.40$). 

\subsection{Ranking Proposals Analysis}

In this section, we analyze the ranking proposals methods with the estimates from \emph{GAN-OURS}. Tab.~\ref{tab:acc_rank_proposals} shows the accuracy of the ranking proposals methods in the test sets 
of the train test split settings (TH\"{O}R and Argoverse).
It is evident that the accuracy of the ranking proposals methods directly
affects both Top-3 and Top-1 results presented in Tab.~\ref{tab:res1_adefde} and Tab.~\ref{tab:res2_t1adefde}, respectively. 
A broad view of the results shows that the more accurate our ranking proposals methods are, the better the ADE/FDE scores, specially Top-1 ADE/FDE. 
As it is possible to observe within each clustering method and dataset, \emph{anet} is (statistically) better only in TH\"{O}R for 
all clustering methods. We speculate that this may be because the clusters in TH\"{O}R are closer to each other as people were moving 
in the same environment (with similar moving patterns) in both train and test sets. This phenomenon is less noticeable in 
Argoverse since different behaviors stem from cars ({\em agents} and {\em av}) and other road agents ({\em others}). Similarly, in the 
leave-one-dataset-out setting (see Tab.~\ref{tab:acc_rank_proposals_loo}), the diversity of behaviors come from the different datasets. In the majority of these datasets, our proposed distance-based 
methods (\emph{cent}, \emph{neigh-ds} and \emph{neigh-fs}) provide better results than an auxiliary deep neural network (\emph{anet}) across the different clustering methods.
Furthermore, while a constant-width MLP-based \emph{anet} provides accurate estimates, it computationally scales 
as $\mathcal{O}({N_L} H_U^2)$, where $N_L$ is the number of layers of the MLP network and $H_U$ is the number of hidden units per layer.
On the contrary, distance-based methods, \emph{centroids} and \emph{neighbors}, require $\mathcal{O}(K)$ and $\mathcal{O}(K N_{neig})$ computations, respectively.
Hence, besides the fact that distance-based methods do not require training an additional neural network, during inference, they provide similar performance in \emph{linear time} while auxiliary networks require \emph{quadratic time}.

\begin{table}[t]
  \centering
  \caption{Accuracy ($\uparrow$) of the ranking proposals methods in the test sets of the tran test split setting.}
  \label{tab:acc_rank_proposals}
  \resizebox{0.75\columnwidth}{!}{%
  \begin{tabular}{cc|lllllll@{}}
  \toprule
  \begin{tabular}[c]{@{}c@{}}Clustering \\ method\end{tabular} &
    \begin{tabular}[c]{@{}c@{}}Ranking\\ proposals\end{tabular} &
    \multicolumn{1}{c}{TH\"{O}R} &
    \multicolumn{1}{c}{Argoverse} \\ \hline
  \multirow{3}{*}{\emph{k-Means}} &
    \emph{cent}& $83.0\pm0.5$
     & $95.1\pm0.2$
     \\
   &
    \begin{tabular}[c]{@{}c@{}}\emph{neigh-ds}\end{tabular} & $81.1\pm0.8$
     & $94.0\pm0.2$
     \\
   &
    \emph{anet}&
    \multicolumn{1}{c}{$\mathbf{86.6\pm0.6}$} &
    \multicolumn{1}{c}{$\mathbf{95.2\pm0.2}$}  \\ \hline
  \multirow{3}{*}{\emph{TS k-Means}} &
    \emph{cent}& 
    \multicolumn{1}{c}{$66.4\pm0.4$} &
    \multicolumn{1}{c}{$93.9\pm0.9$}  \\
   &
    \begin{tabular}[c]{@{}c@{}}\emph{neigh-ds}\end{tabular} &
    \multicolumn{1}{c}{$65.8\pm0.4$} &
    \multicolumn{1}{c}{$94.6\pm0.3$} \\
   &
    \emph{anet}& $\mathbf{72.7\pm0.8}$
     & $\mathbf{96.0\pm0.1}$
     \\ \hline
  \multirow{4}{*}{\emph{FP SC-GAN}} & 
    \emph{cent}& $64.3\pm5.6$
     & $\mathbf{96.5\pm0.4}$
     \\
   &
   \begin{tabular}[c]{@{}c@{}}\emph{neigh-fs}\end{tabular} & $68.0\pm2.7$
     & $96.3\pm5.1$
     \\
   &
    \emph{anet}& $\mathbf{70.3\pm1.3}$
     & $94.9\pm0.7$
     \\ \bottomrule
    
  \end{tabular}%
  }
  \end{table}

\begin{table}[t]
  \centering
  \caption{Accuracy ($\uparrow$) of the ranking proposals methods in the test sets in the 
leave-one-dataset-out setting.}
  \label{tab:acc_rank_proposals_loo}
  \resizebox{\columnwidth}{!}{%
  \begin{tabular}{cc|lllllll@{}}
  \toprule
  \begin{tabular}[c]{@{}c@{}}Clustering \\ method\end{tabular} &
    \begin{tabular}[c]{@{}c@{}}Ranking\\ proposals\end{tabular} &
    \multicolumn{1}{c}{ETH} &
    \multicolumn{1}{c}{HOTEL} &
    \multicolumn{1}{c}{UNIV} &
    \multicolumn{1}{c}{ZARA1} &
    \multicolumn{1}{c}{ZARA2} \\ \hline
  \multirow{3}{*}{\emph{k-Means}} &
    \emph{cent}
     & $94.9\pm1.0$
     & $\mathbf{100.0\pm0.0}$
     & $\mathbf{84.0\pm0.4}$
     & $92.8\pm0.2$
     & $94.7\pm0.3$
     \\
   &
    \begin{tabular}[c]{@{}c@{}}\emph{neigh-ds}\end{tabular} 
     & $\mathbf{97.7\pm0.8}$
     & $99.2\pm0.3$
     & $80.8\pm1.0$
     & $89.7\pm0.9$
     & $\mathbf{95.5\pm0.2}$
     \\
   &
    \emph{anet}&
    \multicolumn{1}{c}{$93.3\pm2.0$} &
    \multicolumn{1}{c}{$98.6\pm0.4$} &
    \multicolumn{1}{c}{$82.5\pm0.9$} &
    \multicolumn{1}{c}{$\mathbf{93.9\pm0.9}$} &
    \multicolumn{1}{c}{$95.2\pm0.4$} \\ \hline
  \multirow{3}{*}{\emph{TS k-Means}} &
    \emph{cent}& 
    \multicolumn{1}{c}{$95.3\pm1.0$} &
    \multicolumn{1}{c}{$\mathbf{100.0\pm0.0}$} &
    \multicolumn{1}{c}{$\mathbf{86.6\pm0.2}$} &
    \multicolumn{1}{c}{$92.5\pm0.2$} &
    \multicolumn{1}{c}{$94.5\pm0.1$} \\
   &
    \begin{tabular}[c]{@{}c@{}}\emph{neigh-ds}\end{tabular} &
    \multicolumn{1}{c}{$\mathbf{97.7\pm0.8}$} &
    \multicolumn{1}{c}{$99.3\pm0.0$} &
    \multicolumn{1}{c}{$82.0\pm0.9$} &
    \multicolumn{1}{c}{$90.6\pm0.6$} &
    \multicolumn{1}{c}{$94.9\pm0.1$} \\
   &
    \emph{anet}
     & $94.5\pm0.8$
     & $98.3\pm0.6$
     & $84.1\pm0.4$
     & $\mathbf{93.2\pm0.6}$
     & $\mathbf{95.0\pm0.3}$
     \\ \hline
  \multirow{4}{*}{\emph{FP SC-GAN}} & 
    \emph{cent}
     & $70.2\pm11.2$
     & $78.8\pm8.7$
     & $64.9\pm2.1$
     & $\mathbf{90.6\pm3.5}$
     & $\mathbf{91.4\pm4.0}$
     \\
   &
   \begin{tabular}[c]{@{}c@{}}\emph{neigh-fs}\end{tabular} 
     & $\mathbf{75.3\pm5.3}$
     & $\mathbf{90.6\pm6.2}$
     & $\mathbf{65.2\pm2.8}$
     & $89.7\pm5.1$
     & $89.2\pm2.2$
     \\
   &
    \emph{anet}
     & $68.6\pm8.0$
     & $83.3\pm7.9$
     & $55.8\pm2.6$
     & $\mathbf{90.6\pm2.7}$
     & $90.2\pm3.2$
     \\ \bottomrule
    
  \end{tabular}%
  }
  \end{table}
\section{Conclusion}

In this paper, we propose a multi-stage probabilistic trajectory predictor: displacement space 
transformation, clustering stage, proposals generation, and ranking proposals.
At the clustering stage, we propose a novel trajectory clustering method (FP SC-GAN) based on deep features from a 
GAN framework. We also propose distance-based ranking proposals methods for assigning probabilities to the predictions.

We test our methods in two settings: train-test split and leave-one-dataset-out. The former 
comprise human and road agents' trajectory data, while the latter only address pedestrians' trajectory data. 
Experimental results show that our system surpasses the multimodal generative baselines (with \emph{k-variety loss})
in Top-3 ADE/FDE scores. It also performs equally or better than context-free generative baselines in Top-1 ADE/FDE scores. Moreover, the proposed 
clustering method copes better with distributional shifts (HOTEL dataset) than traditional clustering methods.
Finally, our ranking proposals methods based on distance similarity measures perform globally better than auxiliary deep neural 
networks. It is even more remarkable since these mechanisms do not require training an additional neural network and 
run in \emph{linear time}, while a constant-width MLP runs in \emph{quadratic time}.


\bibliographystyle{IEEEtran}
\bibliography{refs.bib}

\end{document}